\documentclass[10pt,conference]{IEEEtran}
\IEEEoverridecommandlockouts
\usepackage{graphicx}
\usepackage{booktabs}
\usepackage{xurl}
\usepackage{listings}
\usepackage{xcolor}
\usepackage{xspace}
\usepackage{microtype}

\definecolor{codebg}{HTML}{F8FAFC}
\definecolor{codeframe}{HTML}{CBD5E1}
\definecolor{codekw}{HTML}{1E3A8A}
\definecolor{codecmt}{HTML}{64748B}

\newcommand{\tool}{LGMT\xspace}

\begin{document}
\bstctlcite{IEEEexample:BSTcontrol}

\title{\tool: A Tool for Testing the Logical Reasoning Reliability of LLMs}

\author{
\IEEEauthorblockN{Zenghui Zhou, Xiaoke Fang, Zheng Zheng}
\IEEEauthorblockA{School of Automation Science and Electrical Engineering, Beihang University\\
Xueyuan Road, Haidian District, Beijing, 100191, China\\
\{zhouzenghui, fangxk, zhengz\}@buaa.edu.cn}
}
\maketitle

\begin{abstract}
Large Language Models (LLMs) increasingly serve as reasoning components in modern software systems, yet a model that answers a logical query correctly can contradict itself when the same problem is restated in an equivalent form. Static benchmarks evaluate queries in isolation and therefore fail to expose such inconsistencies. To address this limitation, we present \tool, an open-source tool that detects reasoning inconsistencies through metamorphic relations grounded in first-order logic. Given a source test case in premise-conclusion form, \tool generates logically equivalent follow-up cases and checks whether the model's entailment predictions agree across the pair, without ground-truth labels. The tool provides 20 built-in metamorphic relations, supports custom ones verified by the Z3 theorem prover, and runs from a CLI or a web studio, including an offline mock mode requiring no API key. Across six widely used LLMs on a bundled benchmark of 3,091 metamorphic groups, \tool reveals inconsistency rates ranging from 20.84\% to 29.44\%. \tool is publicly available on GitHub at \url{https://github.com/MaxwelsDonc/lgmt-tool}, with a demonstration video at \url{https://youtu.be/gou4opptu1U}.
\end{abstract}

\section{Introduction}
Software systems increasingly rely on large language models (LLMs) to derive conclusions from domain rules, policies, and constraints~\cite{pan2023logiclm}. For such components, answering isolated benchmark queries correctly is necessary but not sufficient: if a model reverses its judgment when the same problem is phrased in another logically equivalent way, its reasoning cannot be trusted in production. Standard reference-based benchmarks~\cite{han2024folio,parmar2024logicbench} score each query independently against a static label and therefore remain blind to self-contradictions across equivalent formulations.

Metamorphic testing (MT) overcomes this limitation without relying on ground-truth labels: instead of checking individual outputs against an explicit oracle, it verifies whether relations among multiple executions, known as metamorphic relations (MRs), hold. In this paradigm, the original inputs are \emph{source test cases}. Transforming them under an MR yields \emph{follow-up test cases}, and together they form \emph{metamorphic groups}. If a model answers the follow-up test case differently from the source test case under a logic-preserving MR, the divergence exposes a reasoning defect~\cite{zhou2026lgmt}.

Translating this principle into a practical tool requires addressing four concrete engineering tasks: rewriting formal logical structures rather than surface text, realizing only the modified statements in natural language, executing paired queries reliably across thousands of metamorphic groups and heterogeneous model APIs, and tracing aggregate violation rates back to individual cases for root-cause diagnosis. Existing reasoning test generators such as LogicAsker~\cite{wan2024logicasker} and NLP-oriented MT frameworks~\cite{li2024drowzee,wu2025detecting} address only isolated stages. None provides an extensible, end-to-end tool.

We present \tool, an open-source tool that addresses this gap. Practitioners can evaluate any candidate model through OpenAI-compatible endpoints, QA teams can supply custom domain rules in natural language or NL+FOL pairs, and researchers can register custom Python MRs that Z3 verifies for equivalence upon loading~\cite{demoura2008z3}.

This paper makes three primary contributions.
\begin{itemize}
  \item \textbf{End-to-end implementation.} Cases are kept in paired
  natural-language (NL) and first-order-logic (FOL) forms. The tool rewrites the
  FOL with 20 built-in MRs, realizes only the changed statements in NL, queries
  the model on both cases, and reports each disagreement with its evidence.
  \item \textbf{Extensibility.} Custom MRs are Z3-verified at registration.
  \tool{} runs from a CLI or Web Studio, ships as a Docker image, and works
  offline with a mock model and no API key.
  \item \textbf{Bundled benchmark.} 3,091 metamorphic groups derived from FOLIO,
  LogicNLI, and ProVerQA, with results for six LLMs. The full empirical study is
  in the companion paper~\cite{zhou2026lgmt}.
\end{itemize}

\begin{figure*}[t]
\centering
\includegraphics[width=0.84\textwidth]{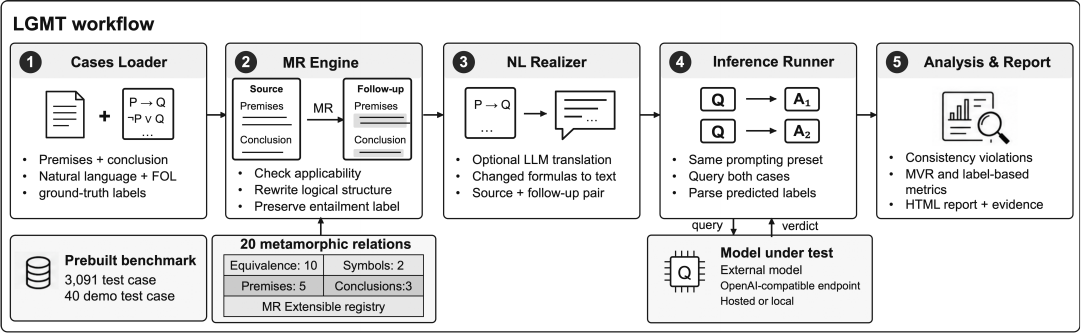}
\caption{The \tool workflow. Five stages turn source test cases into paired queries: one of 20 MRs rewrites the logic, only the changed statements are realized in natural language, and both cases are queried and compared. The tool ships a 3,091-case benchmark (plus a 40-case demo subset) and targets any OpenAI-compatible model.}
\label{fig:arch}
\end{figure*}

\section{Background and Related Tools}
\label{sec:background}

\textbf{Metamorphic Testing.} In \tool, MT is instantiated for three-valued entailment: each case $x=\langle\Gamma,q\rangle$ pairs premises $\Gamma$ with a conclusion $q$, and a model $f$ predicts \textit{True}, \textit{False}, or \textit{Unknown}~\cite{chen1998metamorphic,chen2019metamorphic}. Every MR rewrites a case into a logically equivalent one, so the expected relation is $f(x_s)=f(x_f)$, and any disagreement exposes a defect without labels. \tool implements 20 such MRs across four levels: formula equivalence (MR-E, 10), symbol renaming (MR-S, 2), premise restructuring (MR-P, 5), and conclusion rewriting (MR-C, 3). The companion paper defines them formally and establishes the completeness of the MR-E1 normalization subset~\cite{zhou2026lgmt}.

\textbf{Related Tools.} Prior tools address adjacent parts of the problem space.
Reasoning test generators such as LogicAsker~\cite{wan2024logicasker} synthesize tests from fixed logic templates rather than transforming user-supplied natural-language cases.
Neurosymbolic pipelines such as Logic-LM~\cite{pan2023logiclm} and LINC~\cite{olausson2023linc} translate natural language into FOL so an external solver can compute the answer, evaluating the translation-plus-solver pipeline rather than testing whether the LLM itself reasons consistently.
By contrast, general-purpose MT frameworks such as GeMTest~\cite{speth2025gemtest} orchestrate Python MRs in \texttt{pytest} but provide no built-in FOL support, while NLP-focused MT tools~\cite{li2024drowzee,wu2025detecting,yang2025hallucinationa,cho2025metamorphic} perturb inputs lexically or vary execution paths, so their transformations are not guaranteed to preserve semantics.
\tool combines symbolic rigor with natural-language evaluation by performing Z3-checked rewrites on FOL syntax trees and selectively realizing only the modified statements into natural language.

\section{\tool: Architecture and Implementation}
\label{sec:tool}

Figure~\ref{fig:arch} illustrates the five stages that implement the engineering tasks outlined above.
Below we trace how a test case flows through the pipeline, summarizing foundational concepts from our companion study~\cite{zhou2026lgmt} while detailing the tool's design, extensibility, and runtime mechanisms.

\subsection{Architecture and Case Representation}

Transformations must operate on unambiguous formal syntax, whereas LLMs must be evaluated in natural language. To decouple these layers, the \textbf{Cases Loader} (Stage 1 in Figure~\ref{fig:arch}) represents every record as a dual \texttt{TestCase} pairing natural-language text with FOL formulas. If only text is provided, the loader invokes an LLM translator to synthesize and NLTK-validate the formulas. If only FOL is supplied, the formulas are copied as text.

The case then advances through four decoupled stages: the \textbf{MR Engine} (\texttt{lgmt} \texttt{build-mgs}) rewrites FOL syntax trees, the \textbf{NL Realizer} (inside \texttt{build-mgs}, audited via \texttt{lgmt} \texttt{audit}) renders modified formulas back into text, the \textbf{Inference Runner} (\texttt{lgmt} \texttt{run}) queries the model on the pair, and the \textbf{Analysis \& Report} (\texttt{lgmt} \texttt{report}) checks prediction consistency and compiles the HTML report. Stages communicate via JSONL files, so users can run the pipeline end-to-end or invoke commands individually.

\subsection{Pluggable MR Engine with Z3-Backed Checks}

With formal structure separated from surface phrasing, the \textbf{MR Engine} (Stage 2) parses each FOL formula into an abstract syntax tree via \texttt{nltk.sem.logic} and applies tree-level rewrites across the 20 built-in MRs (e.g., De Morgan under MR-E, predicate/constant renaming under MR-S, premise reordering under MR-P, and conclusion double negation under MR-C). Because rewrites act strictly on syntax trees, untouched premises remain verbatim and the engine records exactly which statements were altered.

\par\addvspace{2pt}%
\refstepcounter{lstlisting}\label{lst:mr}%
{\centering\footnotesize\lstlistingname~\thelstlisting. Custom MR, whose registration checks equivalence with Z3.\par}%
\vspace{2pt}
\begin{lstlisting}[language=Python]
import copy
from nltk.sem.logic import LogicParser, NegatedExpression
from lgmt.core.structure import (MetamorphicRelation, TestCase)

class MRCxDoubleNegation(MetamorphicRelation):
    def __init__(self):
        self.index, self.name = "MR-CX", "Double Negation"
    def source_templates(self) -> list[str]:
        return []
    def check_applicability(self, tc: TestCase) -> bool:
        return bool(tc.conclusion_logic)
    def transform(self, tc: TestCase) -> TestCase:
        follow = copy.deepcopy(tc)
        ast = LogicParser().parse(tc.conclusion_logic)
        follow.conclusion_logic = str(
            NegatedExpression(NegatedExpression(ast)))
        follow.conclusion_text = (
            "[Changed_conclusion] It is not the case "
            f"that not ({tc.conclusion_text})")
        return follow
\end{lstlisting}

\textbf{Custom-MR Registration.} To target domain-specific patterns, the engine exposes a pluggable Python interface (\texttt{MetamorphicRelation}, Listing~\ref{lst:mr}) loadable via \texttt{--mr-file} or the Web Studio. Because hand-written AST manipulations can alter truth conditions, registration runs \texttt{lgmt.mrs.verifier} before any model API call: it applies the candidate MR to five canonical FOL samples and queries Z3 to confirm each follow-up remains equivalent to its source, rejecting the MR if any applicable sample fails or none applies.

\begin{figure*}[t]
\centering
\includegraphics[width=0.84\textwidth]{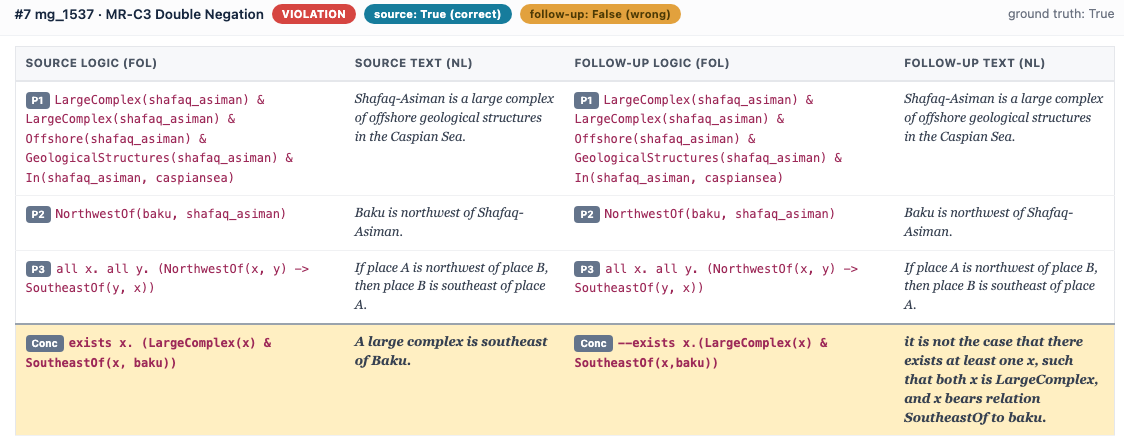}
\caption{A violation card for MR-C3 in \tool's HTML report. The card shows the four-column FOL/NL diff and the side-by-side reasoning traces that localize where and why the model's prediction diverges.}
\label{fig:report}
\end{figure*}

\subsection{Selective Realization and Fidelity Auditing}

Even when Z3 guarantees equivalence of the follow-up FOL skeleton, re-translating the whole case via an LLM risks paraphrasing unchanged premises and introducing semantic drift. To confine generation to the rewritten logic, each MR prefixes only the modified statements with a marker (\texttt{[Changed\_0]} or \texttt{[Changed\_\allowbreak conclusion]}). The \textbf{NL Realizer} (Stage 3) translates only those tagged statements under a constrained prompt that forbids simplification, quantifier reordering, or extra deductions, keeping all other sentences verbatim.

To verify that these localized translations preserve formula semantics, \texttt{lgmt} \texttt{audit} runs an independent judge model that compares each tagged FOL statement against its realized sentence and flags mismatches.
\subsection{Paired Inference and Diagnostic Reporting}

\textbf{Reasoning-Aware Inference Runner.} Given a faithful pair $\langle x_s, x_f \rangle$, the \textbf{Inference Runner} (Stage 4) queries the model concurrently under one of four prompting presets (\texttt{zero\_\allowbreak direct}, \texttt{zero\_\allowbreak cot}, \texttt{few\_\allowbreak direct}, \texttt{few\_\allowbreak cot}), repairs malformed JSON, validates outputs against a strict schema, and supports checkpoint resumption (\texttt{--resume}). Unparseable responses are marked \texttt{is\_valid=false} so they cannot skew metrics. Reasoning models (e.g., o4-mini, DeepSeek Reasoner) add protocol friction: they reject sampling parameters, often emit answers only in \texttt{reasoning\_\allowbreak content}, and hit token caps during long chains of thought. The runner omits rejected parameters, falls back to \texttt{reasoning\_\allowbreak content}, and escalates the token budget from 8,192 to 65,536 when \texttt{finish\_\allowbreak reason} is \texttt{"length"}.

\textbf{Analysis and Metrics.} In Stage 5, the analysis stage maps both predictions to \textit{True}, \textit{False}, or \textit{Unknown} and checks whether $f(x_s)=f(x_f)$. \texttt{lgmt} \texttt{report} then quantifies reliability from two perspectives: the \emph{metamorphic violation rate (MVR)}, the share of groups with $f(x_s)\neq f(x_f)$, measures overall self-contradiction. The \emph{hidden defect rate (HDR)}, the share of groups whose source is correct but follow-up is not, captures the drop from static to consistent accuracy. The \emph{false unreported rate (FUR)} is the share of groups where both answers are wrong but agree.

\textbf{Three-Layer HTML Report.} Because summary metrics say how often a model fails but not where or why, \texttt{lgmt} \texttt{report} compiles a self-contained HTML file (Figure~\ref{fig:report}) as a progressive drill-down: (1) a leaderboard calibrated by dataset and prompting strategy, (2) category- and MR-level tables isolating the most failure-prone transformations, and (3) expandable violation cards, reached by clicking any violation count, showing the four-column FOL/NL diff and side-by-side reasoning traces.

\subsection{Implementation, Packaging, and Web Studio}

To lower the barrier to adoption across environments, \tool is an MIT-licensed Python distribution (with a Dockerfile) providing a seven-command CLI (\texttt{demo}, \texttt{build-mgs}, \texttt{audit}, \texttt{run}, \texttt{report}, \texttt{list-mrs}, \texttt{web}) and a zero-dependency Web Studio. A deterministic \texttt{MockLLMClient} with two behavior modes, \texttt{consistent} ($\mathrm{MVR}=0\%$) and \texttt{flawed} (injecting reasoning defects), lets every workflow run offline without API keys. Together with Z3 equivalence checks across all 20 built-in MRs, it powers a 147-test \texttt{pytest} suite on Linux, macOS, and Windows under Python 3.10 and 3.13.

\section{Tool Usage and Workflows}
\label{sec:usage}

The CLI and Web Studio organize \tool around three progressive workflows.

\textbf{Workflow 1: Offline Tour.} A new user or reviewer can run the full pipeline immediately after installation (\texttt{pip install -e .} or \texttt{docker build -t lgmt .}), without an API key or network:
\begin{lstlisting}
lgmt demo; lgmt web --port 7860
\end{lstlisting}
\texttt{lgmt demo} evaluates 40 bundled FOLIO-derived groups against the flawed mock mode and writes \texttt{demo\_output/report.html} in under 30 seconds. \texttt{lgmt demo --full} exercises all 3,091 groups offline.

\textbf{Workflow 2: Pre-Deployment Benchmarking.} To evaluate a real model, a team points \tool at any OpenAI-compatible endpoint (a local vLLM or Ollama server, or a commercial API) and runs:
\begin{lstlisting}
lgmt run --mgs benchmark --model llama-3.3-70b --output res.jsonl
lgmt report --results res.jsonl --output report.html
\end{lstlisting}
The \texttt{benchmark} alias resolves to the 3,091-group set (6,182 requests per model, $\sim$$10^7$ tokens), whereas \texttt{demo} resolves to the 40-group subset (80 requests), an inexpensive check of credentials and output parsing before a full run.

\textbf{Workflow 3: Custom Cases, Custom MRs, and CI Integration.} Teams supply cases as plain English, NL+FOL pairs, or JSONL. \tool synthesizes and validates any missing FOL, applies built-in or Z3-verified custom MRs, and tests the target model:
\begin{lstlisting}
lgmt build-mgs --input cases.jsonl --mr-file my_mr.py --model gpt-5.2 --output my_mgs.jsonl
lgmt run --mgs my_mgs.jsonl --model gpt-5.2 --output res.jsonl
\end{lstlisting}
The output feeds the same HTML report, which CI/CD jobs can use to gate model upgrades whenever MVR or HDR exceeds a threshold.

\begin{table}[t]
\centering
\caption{Validation on the 3,091-Group Benchmark (Zero-Shot CoT). All six models exceed a 20\% MVR, indicating substantial reasoning inconsistency that static accuracy misses.}
\label{tab:validation}
\footnotesize
\begin{tabular}{lcccc}
\toprule
Model & Acc$_\mathrm{static}$ & Acc$_\mathrm{cons}$ & MVR & HDR \\
\midrule
GPT-5.2           & 73.34\% & 56.84\% & 26.76\% & 16.50\% \\
Claude 4.5 Sonnet & 73.53\% & 60.97\% & 20.84\% & 12.56\% \\
Llama 3.3 70B     & 65.38\% & 52.38\% & 25.43\% & 13.01\% \\
DeepSeek Chat     & 69.91\% & 55.65\% & 24.72\% & 14.27\% \\
o4-mini           & 69.36\% & 56.20\% & 22.06\% & 13.17\% \\
DeepSeek Reasoner & 72.44\% & 56.71\% & 29.44\% & 15.72\% \\
\bottomrule
\end{tabular}
\end{table}

\section{Validation}
\label{sec:validation}

To demonstrate the diagnostic insights \tool provides in practice, we ran the full pipeline on the bundled benchmark of 3,091 metamorphic groups derived from FOLIO~\cite{han2024folio} (2,673), LogicNLI~\cite{tian2021diagnosing} (93), and ProverQA~\cite{qi2025large} (325) across six LLMs and four prompting strategies. Table~\ref{tab:validation} summarizes the Zero-shot CoT results, and the companion paper reports the complete study~\cite{zhou2026lgmt}.

As Table~\ref{tab:validation} shows, every evaluated model has substantial reasoning inconsistency, with MVRs ranging from 20.84\% (Claude 4.5 Sonnet) to 29.44\% (DeepSeek Reasoner). Requiring consistency across equivalent follow-up cases drops accuracy from 65.38\%--73.53\% ($\mathrm{Acc}_\mathrm{static}$) to 52.38\%--60.97\% ($\mathrm{Acc}_\mathrm{cons}$), with HDRs of 12.56\%--16.50\% that single-query evaluation misses. Breaking results down by MR category shows that failures concentrate in specific structural operations: conclusion rewriting (MR-C) and symbol renaming (MR-S) trigger the highest aggregate MVRs (33.78\% and 33.06\%), whereas premise restructuring (MR-P) triggers the lowest (16.33\%). This illustrates how \tool pinpoints concrete reasoning vulnerabilities beyond static accuracy.

\section{Conclusion}
\label{sec:conclusion}

We presented \tool, an open-source tool that operationalizes logic-grounded metamorphic testing into an end-to-end workflow. Coupling FOL syntax-tree rewrites and Z3-verified custom MRs with selective realization, reasoning-aware inference, and drill-down HTML reports, \tool enables practitioners to audit LLM reasoning consistency from the CLI or Web Studio without ground-truth labels. While realizing rewritten statements still relies on an LLM and the current engine targets FOL entailment, extending it to composed MRs and broader formalisms is a natural next step. The tool, documentation, and benchmark are available at \url{https://github.com/MaxwelsDonc/lgmt-tool}.

\section*{Acknowledgment}
Generative AI tools were used to assist with code refactoring and manuscript editing. All content was reviewed and verified by the authors, who take full responsibility for the work.

\bibliographystyle{IEEEtran}
\bibliography{refs}

\end{document}